# Executable Ontologies:
# Synthesizing Event Semantics with Dataflow Architecture

Aleksandr Boldachev
boldachev@gmail.com

## Abstract

This paper presents boldsea, Boldachev's semantic-event approach — an architecture for modeling complex dynamic systems using executable ontologies — semantic models that act as dynamic structures, directly controlling process execution. We demonstrate that integrating event semantics with a dataflow architecture addresses the limitations of traditional Business Process Management (BPM) systems and object-oriented semantic technologies. The paper presents the formal BSL (boldsea Semantic Language), including its BNF grammar, and outlines the boldsea-engine's architecture, which directly interprets semantic models as executable algorithms without compilation. It enables the modification of event models at runtime, ensures temporal transparency, and seamlessly merges data and business logic within a unified semantic framework.

## 1. Introduction

### 1.1. Problem Context

Contemporary enterprise systems face a core tension that limits adaptability and scale. BPM systems, for instance, leverage imperative, sequentially-executed models to automate predefined workflows efficiently. However, these models remain disconnected from the underlying data structures — particularly semantic ones — resulting in rigid schemas that are difficult to modify. Similarly, event-driven architectures (EDA) such as Event Sourcing, while effectively capturing change histories as event streams, typically treat events as unstructured 'raw' data, which hinders validation and system integration. Meanwhile, object-oriented semantic technologies such as RDF/OWL shine in depicting domain knowledge as graphs but were not designed to model dynamic activities or drive business process execution.

## 1.2. The Concept of Executable Ontologies

We address these limitations by implementing the executable-ontology paradigm. Within this approach, semantic models are not static descriptions but dynamic structures that directly manage process execution.

Building on the *subject–event ontology* (Boldachev, 2015), we model activity as a *temporal graph of semantically typed events* and execute it via an *event-based dataflow* mechanism. Process execution within this graph employs an *event-based version of the dataflow paradigm*, which implements asynchronous event generation upon the fulfillment of logical conditions specified in the semantic models.

This synthesis bridges the semantic gap between process management and knowledge representation. It offers key advantages, including:

- Parallel process execution, managed by data readiness conditions rather than a predefined sequence of steps.
- Complete temporal traceability of all operations for auditing, analytics, and machine learning.
- Modification of semantic models at runtime, without system interruption.
- No-code development of models, reducing the time to create business processes from weeks to days and making them accessible to business analysts without programming skills.
- AI integration, allowing LLMs to generate executable business process models from text descriptions with near real-time automatic validation (Kourani, 2024).

The entire approach is formalized through the *boldsea Semantic Language (BSL)* — a domain-specific language that ensures the system's predictability and verifiability.

## 1.3. Contribution and Structure of the Paper

The goal of this paper is to outline the theoretical and some practical foundations of the technology. The text does not contain a complete description of the event semantics specification and the boldsea-engine architecture (which is currently in a beta version and undergoing stress tests). The paper presents the fundamentals of event semantics (Section 3), explains the principles of dataflow modeling with a practical example (Section 4), introduces the formal model of the BSL language with its BNF grammar for model verification (Section 5), and describes the boldsea-engine architecture (Section 6). The concluding sections discuss the advantages and limitations of the technology and outline future research directions. The Appendix contains a glossary and a brief description of the integrated development environment.

# 2. Related Works and Positioning

Our approach emerges at the intersection of three actively developing areas of information technology: semantic knowledge representation, business process management, and event-driven architecture. An analysis of each of these areas reveals fundamental gaps that prevent the creation of truly flexible, adaptive systems. We propose bridging these gaps through *executable ontologies* methodology.

## 2.1. The Semantic-Process and Structural-Execution Gaps

Standards like RDF, OWL, and SPARQL provide a robust framework for formal knowledge representation, adeptly capturing entities and their static relationships (Berners-Lee, 2001). However, they were not originally designed to model dynamics. While extensions such as OWL-Time and T-SPARQL, or 4D ontologies like ISO 15926, enable temporal tracking of changes, they fall short of rendering ontologies executable. These tools clarify which entities and relations exist, but say little about which actions produced a state or what follows from it.

Conversely, Business Process Management (BPM) systems, standardized in notations such as BPMN, focus exclusively on process automation, decoupling process logic from the semantics of the underlying data. Even advanced approaches, such as YAWL or Petri nets, maintain this fundamental separation. The result is the *semantic-process gap*: knowledge systems cannot manage processes, and process systems operate with semantically poor data (van der Aalst, 2003). Similarly, event-driven architectures (EDA, Event Sourcing), while offering an elegant solution for recording the dynamics of a data stream (Fowler, 2005), in most cases use weakly structured messages as events, which lack semantic typing, again creating a semantic gap. Likewise, the *dataflow paradigm*, while contrasting the rigid control-flow approach with native parallelism and scalability, traditionally focuses on computational results rather than the semantic modeling of business processes. Together, this yields a structural–execution gap: either asynchronous execution without semantics or semantics without execution.

Attempts to combine semantics and business logic are also made by introducing semantic business rules into RDF/OWL ontologies, using specifications like SWRL or SPIN. In this approach, logic is defined as a set of rules executed by an external engine over object graphs. However, such a hybrid architecture maintains a fundamental separation: the ontology remains a passive knowledge store, and the rule engine is a separate, external entity. As a result, despite the addition of dynamics, the semantic model itself does not become inherently executable. This leaves the structural-execution gap unresolved, where knowledge and the logic for processing it exist as separate components of the system.

*Table 1. Comparison with Existing Solutions*

| Aspect | BPM (BPMN) | RDF/OWL | Event Sourcing | boldsea |
|---|---|---|---|---|
| Temporality | Limited | Absent | Present | Native |
| Semantics | Weak | Strong | Absent | Strong |
| Executability | Yes | No | Partial | Full |

## 2.2. Synthesis through Executable Ontologies

The proposed architecture's novelty lies in deliberately synthesizing the strengths of all three approaches to bridge these gaps:

- From semantic technologies, the architecture inherits the representation of knowledge in the form of ontologies based on unified vocabularies, but makes them executable through a *dataflow mechanism* built directly into the model's semantics.
- From BPM systems, the focus on process automation is adopted, but the rigid control-flow logic is replaced with a flexible *event-based dataflow approach*, where the execution sequence is determined by the readiness conditions of the data.
- From event-driven architectures, temporality and the atomicity of data recording are adopted, but the sequence of events is not set externally; instead, it is generated by the execution of the semantic models on the stream itself, and the semantic typing of events makes them understandable for machine interpretation and validation.

Thus, the described architecture is not an evolutionary development of any one of the three directions. Instead, it represents an integration of semantic technologies, BPM, and EDA into a unified approach. In this framework, ontologies evolve beyond static descriptions, business processes gain semantic transparency, and events become foundational elements of executable models that seamlessly link data and processing logic within a single temporal graph.

# 3. Fundamentals of Event Semantics

Event semantics is based on a fundamental rethinking of the nature of activity and the principles of its description. Instead of the traditional separation into static objects and dynamic processes, a single model is proposed in which all activity, including resources and actors, is viewed as a stream of *semantically typed events*. These events, linked by causal relationships, form a temporal graph that allows for the natural modeling of both states and transitions between them within a single executable ontology (Boldachev, 2021).

## 3.1. Event Structure and the Temporal Graph

The event is the core ontological primitive from which all aspects of the modeled activity are built. In contrast to objects, which are temporally extended entities, events are treated as atomic and immutable facts. Each event follows a unified structure:

```
Event = (Id, Base:Type:Value, Actor, Cause, Model, Timestamp)
```

where:

- `Id`: the event identifier,
- `Base: Type: Value`: a semantic triplet that defines the event's content: the *Base* (the entity to which the event applies), the *Type* (the property being recorded), and the *Value* (the specific value of that property); the `Base` field can reference any existing event in the graph, enabling the definition of properties on properties (higher-order properties),
- `Actor`: the identifier of the actor (a person, sensor, software agent) that recorded or initiated the event,
- `Cause`: links to events that causally or logically conditioned this event,
- `Model`: a reference to the model according to which this event was created,
- `Timestamp`: the timestamp.

Example of an event record in JSON format (used for data export/import):

```
{
        "id": "#e398546238c9633c7769c4635e0680154ba2f760",
        "base": "#f8cc0d086cc386b84cbc12a3733789c3dfeea596",
        "type": "Offer",
        "value": "Offer for product turbine for sale",
        "actor": "#01bd7d1f079baabfde314245d7d20062770a499e",
        "model": "Model Processing request",
        "cause": "#f8cc0d086cc386b84cbc12a3733789c3dfeea596",
        "date": "2024-10-23T10:59:29.478Z"
}
```

A key distinction from traditional semantic approaches is the mandatory inclusion of the Actor field, which ensures a subject-centric description, tying every fact to its source. This enables data provenance and allows for tracking the origin of information, as well as capturing discrepancies in observations (from both human actors and sensors).

Furthermore, each event contains the `Cause` field — a set of links to events that served as conditions for its generation (according to the dataflow approach). These causal links form a *directed acyclic graph (DAG)*, in which the nodes are events and the edges are relationships of causal conditioning. This structure naturally allows for:

- recording the native parallelism of actions, since events that are not causally related can occur simultaneously and independently;
- ensuring a complete historical reconstruction, making it possible to restore the state of any individual at any moment in time;

- supporting causal analysis, allowing one to trace which events led to which consequences.

Unlike blockchains with enforced linear ordering, or static RDF graphs that do not record change, the temporal semantic graph natively captures dynamics and causality

## 3.2. Two-Level Data Structure

To ensure the validity and executability of the ontology, a two-level structure of the event graph is used, similar to the TBox/ABox separation in descriptive logics but adapted for dynamic event semantics: events are divided into model events and reification events.

*Model Events* are templates for creating reification events. Grouped into structures, model events form models of concepts or actions, defining the "schema" of the graph. Models define: the semantics of individuals (what properties they can have), restrictions on property values (data types, value ranges, multiplicity), and most importantly, the *conditions* for event execution and access permissions to them. Example of a model (record in BSL notation):

```
Person: Model: Model Person
: Attribute: age
:: Required: 1
:: ValueCondition: $Value >= 0 && $Value <= 120
:: Permission: manager
```

*Reification Events* are specific instances, "facts," created in the course of an activity according to the model event templates.

```
Person: Individual: Smith
: age: 35
```

## 3.3. Basic Principles of Event Semantics

Each event (both model and reification) is anchored to a preceding event, meaning it contains the `Id` of that event in the `Base` field (the `:` symbols denote this event nesting). This principle naturally allows for specifying properties on properties ("temperature is 38 and it's rising") a long-standing challenge in traditional semantics.

In the `Type` field of the semantic triplet, model events can only contain a symbolic identifier of a property (an attribute or relation) described in one of the vocabularies. The use of unified common and industry-specific vocabularies in event semantics, as in other semantic technologies, ensures interoperability, that is, the free exchange of data between independent applications.

Notably, unlike traditional semantic technologies, in event semantics, each reification event is created only and exclusively according to a model event, which is part of the model specified in the `Model` field. This ensures the validation of events according to the conditions and *restricting properties* defined in the models, guaranteeing the semantic correctness of the temporal graph. Furthermore, the unambiguous structuring of the graph by models, i.e., its natural thematic clustering, greatly simplifies data search. The use of event hashes as identifiers for reification events also provides an additional benefit: cryptographic protection of the graph from falsification.

Since each event contains the hash of at least one preceding event in the `Cause` field, changing an event would require rewriting the entire subsequent chain of events in the graph.

Another significant difference between the event ontology and the object ontology is the fundamental separation of the activity modeling level and the classification level. That is, it is accepted that for recording the substantive side of an activity, it is sufficient to operate with reification events without involving Class-Subclass relations. Genus-species relations and other classifications are defined by separate models and are used only for data analysis, display, and search.

# 4. Dataflow Architecture and Executable Semantic Models

The key innovation of the approach is the application of *dataflow architecture* for the execution of semantic models. Unlike traditional BPM systems, which sequentially execute predefined sequences of steps, the event models in the described approach implement an asynchronous principle of control, where individual events are generated independently as logical conditions are met. This approach provides inherent parallelism, facilitates modification, and tightly integrates data and logic.

## 4.1. From Imperative Control-Flow to Event-based Dataflow

Classic BPM systems inherited the imperative control-flow paradigm from the von Neumann architecture, where a process is represented as a graph with explicitly defined transitions:

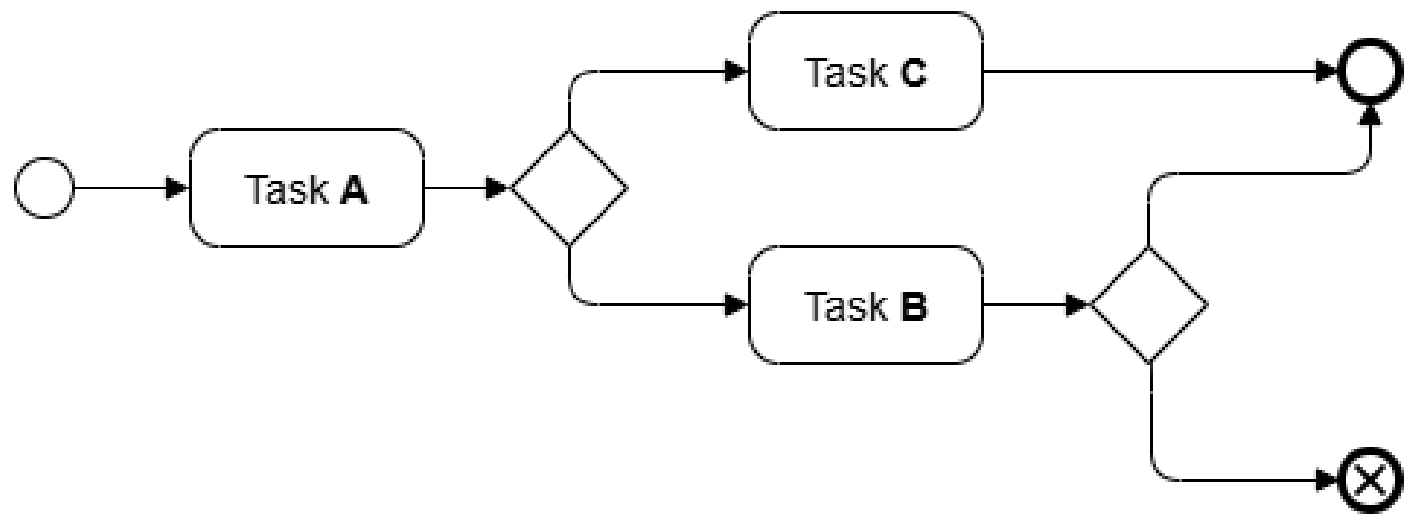

*Figure. 1. Example of a business process model in a BPMN-like notation.*

This approach implies a rigid execution sequence and requires system shutdown to modify the logic.

The dataflow paradigm, described by Dennis (Dennis, 1974), offers an alternative: asynchronous execution of operations as input data becomes available.

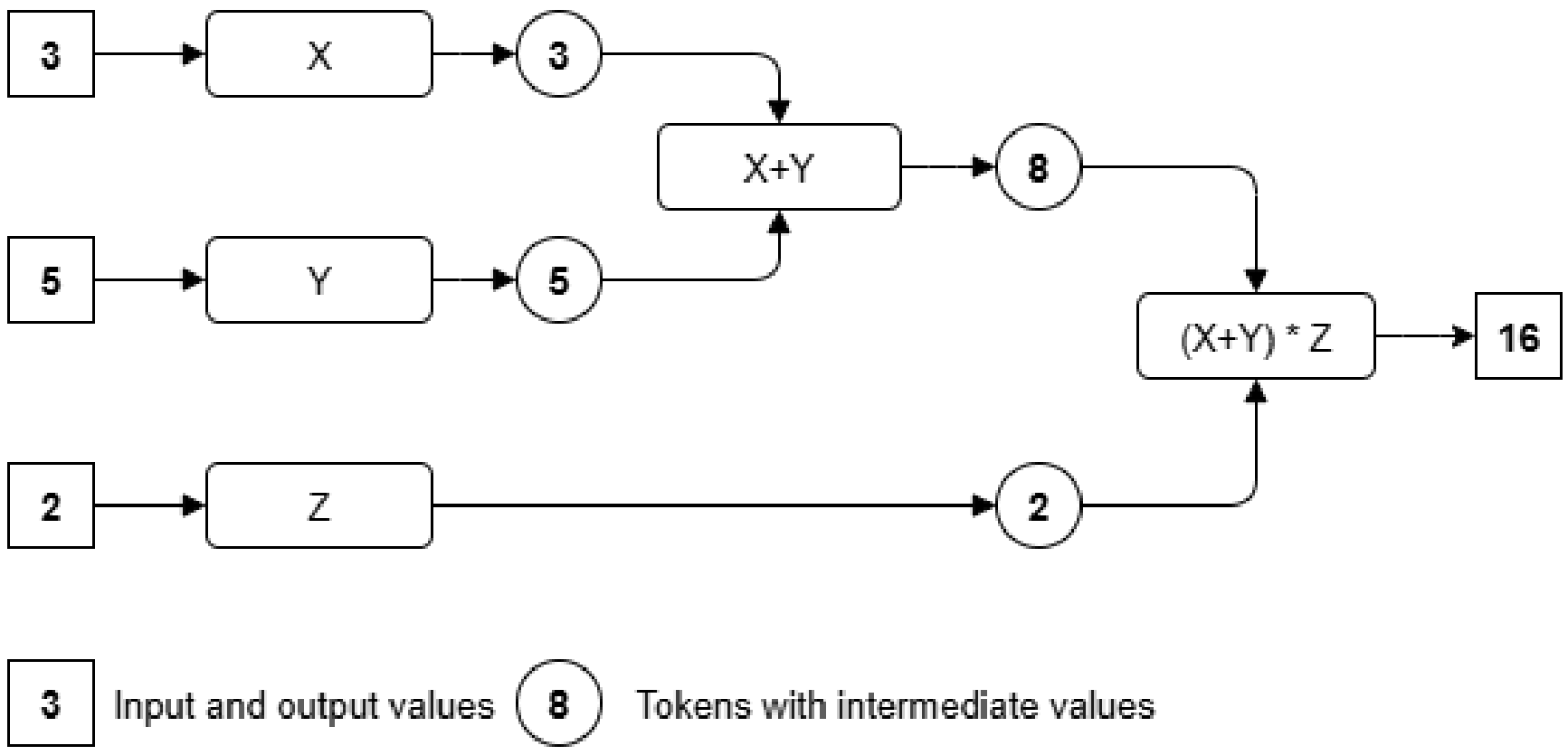

*Figure 2. The original computational version of dataflow architecture: expressions are calculated as data arrives, and the result is passed to the next node.*

The executable ontology architecture adapts this principle for event semantics: *model events* act as "operators," and *reification events* act as "data." A model event is activated — i.e., it generates a new event—not by an external command, but when its logical `Condition` expression evaluates to true as the necessary reification events appear in the graph. This creates a reactive system in which changes cascade through the graph, ensuring native parallelism and eliminating the need for centralized coordination.

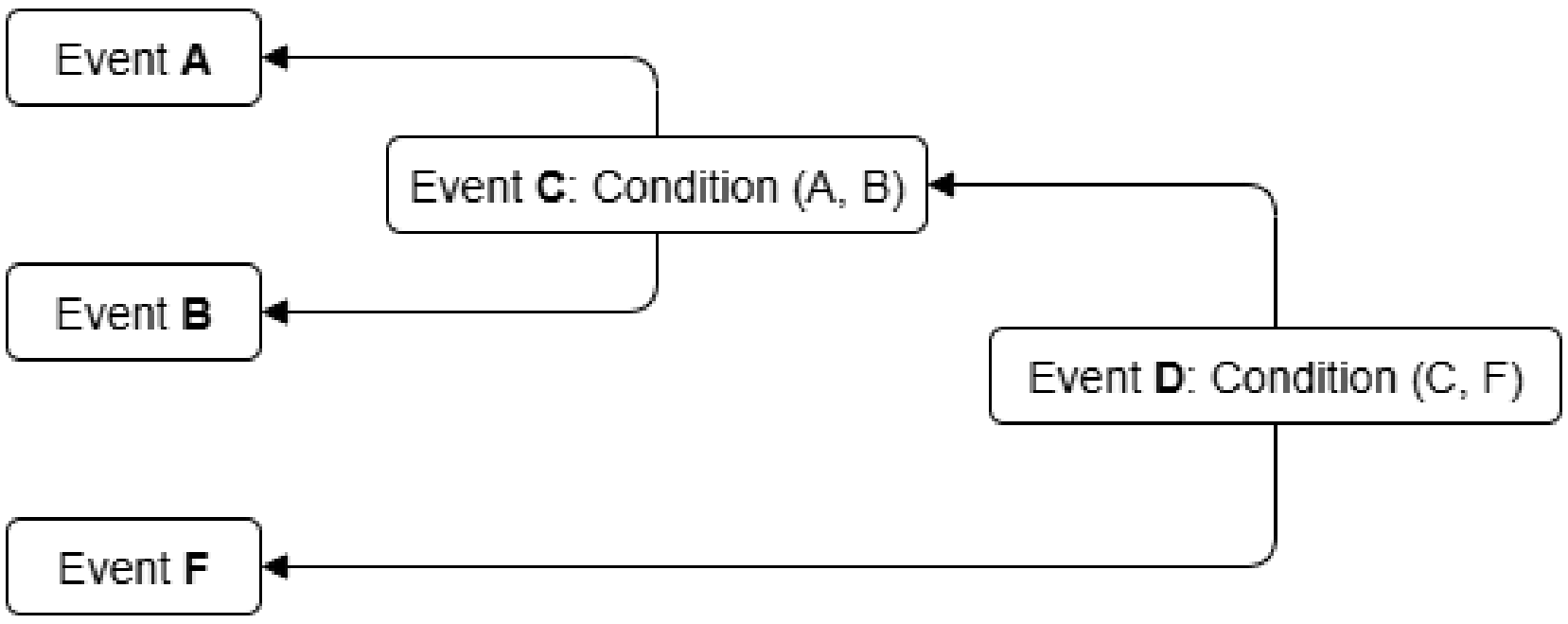

*Figure 3. The event-based version of dataflow architecture: an event is executed when a condition is met. The fundamental difference from the native dataflow architecture is that the result is not passed anywhere, but rather the subscriptions of other events are triggered according to their conditions.*

A significant advantage of the dataflow execution of models is the ability to extend models without changing the functions already implemented. For example, if another event is added to the model shown in the figure, which is triggered by the presence of events B and F, it cannot fundamentally affect the triggering logic of event D.

## 4.2. Declarative Logic Description through Restricting Properties

In event semantics, business logic is expressed not through command sequences but declaratively, using *restricting properties* (`Restrictions`) assigned to model events. It is these constructs that transform a static model into an executable dataflow program.

The basic mechanism that triggers the dataflow logic is the `Condition` property, which is assigned to a model event and contains a logical expression. The truth of this expression is the condition for creating a new reification event based on the model. If the condition is not met when the model is launched, the engine creates an internal subscription and re-evaluates it when relevant data appears in the graph. For example:

```
:: Condition: $.document_verified == true && $.credit_score > 600
```

A model event with this condition will be activated only after the document has been verified and the credit rating exceeds 600. The expressions in `Condition` use a JavaScript-like syntax with extensions for working with the event graph.

The `SetValue` property is used for automatic value assignment. This mechanism primarily ensures the generation of reification events when the condition is met. The value can be a literal, the result of simple calculations, or a combination of complex conditional expressions or graph queries. This allows, for example, dynamically updating the status of a contract based on the actions of the parties:

```
: Attribute: status
:: SetValue: ($.confirmed_by_tenant === true && $.confirmed_by_landlord === true) ?
             "signed" : ($.confirmed_by_tenant == false  ||
             $.confirmed_by_landlord == false ) ? "rejected" : "draft"
```

Security for business logic execution is enforced at the semantic level: each model event can be assigned a `Permission` restricting property, which defines access rights based on roles or dynamic conditions. The value of `Permission` can be a role or an expression/query that returns the identifier of the actor or group of actors who are currently granted access to the model event (`Permission: $($.tenant).actor`). This approach allows for the implementation of complex context-dependent access control schemes.

Finally, `SetDo` allows for initiating system acts, such as creating (`CreateIndividual`) or editing (`EditIndividual`) individuals. This provides event semantics with the ability to manage the full lifecycle of objects in accordance with the rules embedded in the model.

Thus, in the described architecture, the modeling of activity — the creation of a domain ontology and the implementation of business processes on it — is reduced to the declarative description of the conditions and restrictions for each individual model event. The *set* of these rules forms a complex but predictable and verifiable dataflow program that is executed by the boldsea-engine.

In general, the main principle of *executable ontology* can be formulated as follows: *modeling any activity is reduced only and exclusively to defining the conditions for event execution and setting restrictions on their values, that is,* formally, it consists of assigning restricting properties to each model event.

## 4.3. Practical Modeling: Transforming BPMN into an Executable Model

To demonstrate the practical advantages of the event-based dataflow approach, let's consider the process of transforming a traditional BPMN process "Processing Product Request" into an executable semantic model.

The process includes the creation of a request by a `customer`, the formation of an offer by an `employee`, approval by a `manager`, and confirmation by the `customer`.

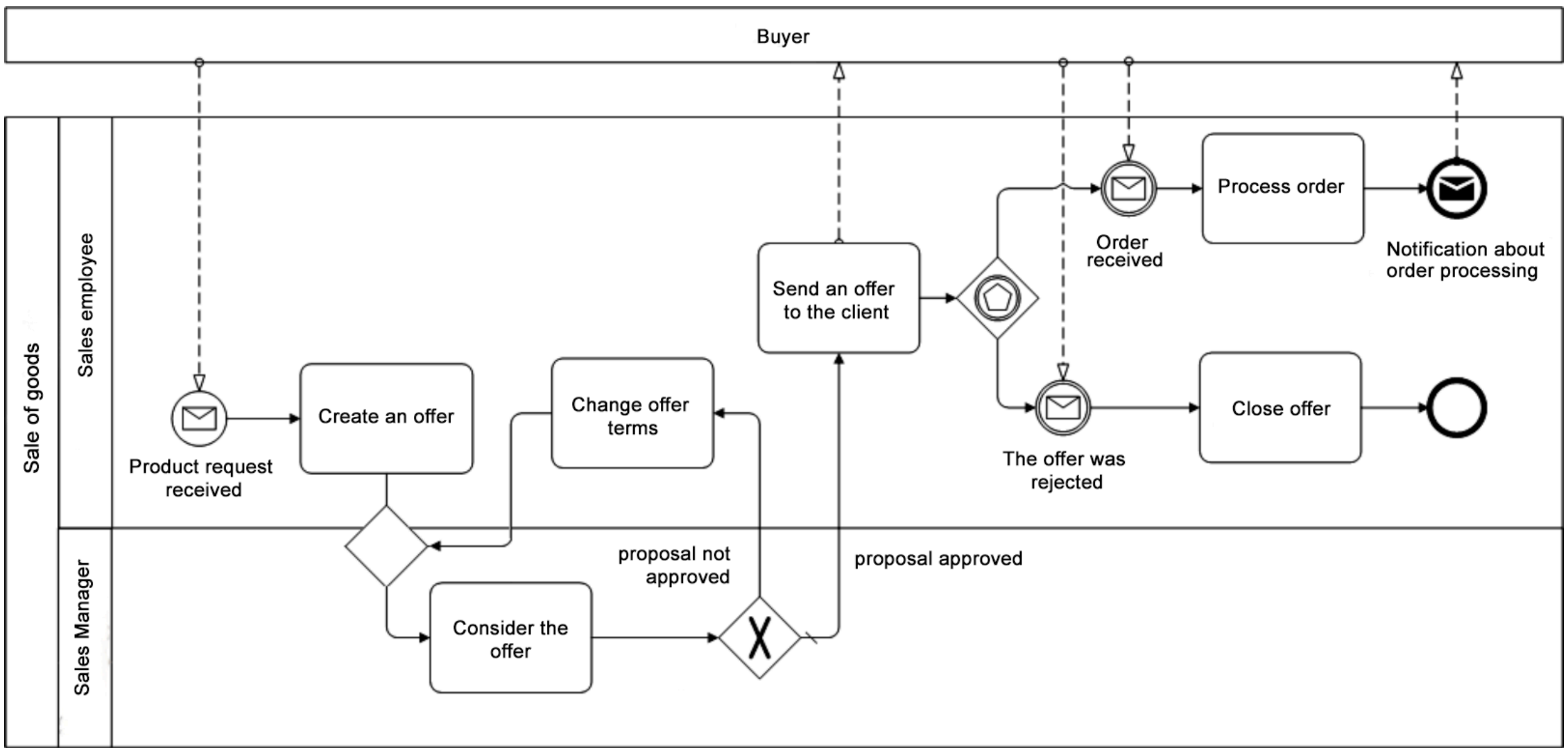

*Figure 4. Example of a business process model in BPMN notation.*

In event semantics, this business process is implemented as a `ProcessingRequest` concept-action model. Instead of gateways, the logic is defined through `Restrictions`:

```
# initiation of concepts, roles and properties is skipped

ProcessingRequest: Model: Model_ProcessingRequest

# The selection of the subject of the request is available to the customer
# until an offer is created

: Relation: subject
:: Permission: customer
:: Condition: $$.offer == undefined

# The employee can edit the offer if it has not yet been approved
# or was returned for revision

: Attribute: offer
:: Permission: employee
:: Condition: ($$.subject != "" && !($$.offer.solution)) ||
              $$.offer.solution == "SendBack"

# The manager makes a decision (approves/rejects)

:: Attribute: solution
::: Permission: manager

# The customer can confirm the order only after the manager's approval
```

```
:: Attribute: confirmation
::: Permission: customer
::: Condition: $.offer.solution == "Accept"

# The status is computed automatically based on all previous events

: Attribute: status
:: SetValue: ($$.Offer.Confirmation === "Yes") ? "process" :
             ($$.Offer.Confirmation === "No" || $$.Offer.Solution === "Reject") ?
             "closed" : undefined

# Creation of individuals/instances of the request processing process
# (actors are not shown)

ProcessingRequest: Individual: PR_001
: subject: Product_A123
: offer: Standard configuration, delivery time 14 days, price 50000 rubles
:: solution: Accept
:: confirmation: Yes
: status: process

# Rejection protocol by manager

ProcessingRequest: Individual: PR_002
: subject: Product_B456
: offer: Special configuration
:: solution: Reject
: status: closed

# Customer refusal protocol

ProcessingRequest: Individual: PR_003
: subject: Product_C789
: offer: Basic configuration
:: solution: Accept
:: confirmation: No
: status: closed
```

This model is declarative (it describes *what* can happen, not *what* must happen), reactive (status updates automatically), and flexible (a new rule, for example, for checking the credit limit, is added as another condition in `Condition` without stopping the process). Each step of the business process is saved in the graph as a reification event with an indication of the actor and the cause, ensuring full traceability.

# 5. Formal Model of Semantics and the BSL Language

The principles of event semantics and dataflow architecture discussed earlier find their specific embodiment in a strict formal model and the *boldsea Semantic Language (BSL)*. BSL is a domain-specific language (DSL) for defining executable ontologies, ensuring predictable, verifiable, and deterministic execution.

## 5.1. Mathematical Foundations and Formal Semantics

BSL is based on a formalism built on first-order logic with temporal extensions similar to Event Calculus (Kowalski, 1986). This ensures mathematical rigor and allows for proving key system properties. The language is based on fundamental axioms (a fragment of the *Formal Semantics Model boldsea* is provided):

### Axiom 1: Structure of Causality and Nesting

An event can only occur if its "context" exists. If a model event `M` is nested under a parent `M_p`, then for each individual `i`, the execution of an event of type `M` requires the existence of an event of type `M_p` for the same `i` has already occurred.

$$\forall i, e, t : (\text{Model}(e) = M \land \text{Indiv}(e) = i \land \text{Happens}(e, t) \land \text{ParentType}(M) = M_p \land n = \text{NestedLevel}(M))$$
$$\implies \exists e_p, t' : (\text{Model}(e_p) = M_p \land \text{Indiv}(e_p) = i \land \text{Cause}(e_p, e) \land \text{Happens}(e_p, t') \land t' < t \land (n-1) = \text{NestedLevel}(M_p))$$

Additional Restriction on Causality:

$$\forall e_1, e_2 : \text{Cause}(e_1, e_2) \implies \text{Time}(e_1) < \text{Time}(e_2)$$

*Restriction:* $n \leq 5$ *for all model events in Boldsea.*

### Axiom 2: Execution Condition (Condition)

The logical `Condition(M)` determines when an event of type `M` can be generated. An event `e` of type `M` for an individual `i` becomes available (`Enabled`) only if all the necessary preceding events `Pred(M)` have already occurred and the condition `Φ_M(i, t)` is true at time `t`.

$$\forall i, t : \left(\bigwedge_{M_p \in \text{Pred}(M)} \exists e_p, t_p : (\text{Model}(e_p) = M_p \land \text{Indiv}(e_p) = i \land \text{Happens}(e_p, t_p) \land t_p < t \land \text{Visible}(e_p)) \land \Phi_M(i, t)\right)$$
$$\implies \text{Enabled}(M, i, t)$$

Here `Enabled(M,i,t)` means that an event of type `M` is available for execution for individual `i` at time `t`. If `Condition` is not specified, `Φ_M(i)` is considered true.

### Axiom 3: Event Generation and Value Assignment (SetValue)

If an event is allowed (`Enabled`), it either occurs automatically or is entered by an actor, provided validation is successful.

- *Automatic Generation (SetValue):* If `SetValue(M)` is an expression specified for `M`, the event is automatically created, and its value is equal to the result of evaluating the function `SetValue(M)`.

$$\text{Enabled}(M, i, t) \land \text{HasSetValue}(M)$$
$$\implies \exists e : (\text{Model}(e) = M \land \text{Indiv}(e) = i \land \text{Happens}(e, t+1) \land \text{Value}(e) = \text{SetValue}(M)(i, t) \land \text{ValidEvent}(e, t+1))$$

The complete axiom set ensures determinism (unique computation results), consistency (absence of contradictions), and termination (finiteness of cascading computations).

## 5.2. BSL Language BNF Grammar

BSL has a formal BNF grammar that ensures the automatic validation of models. Below are fragments of the grammar:

```
# Basic model structure

<model_definition>  ::= <concept_reference> ":" "Model" ":"
                        <model_name> <model_body>
<model_body>        ::= <property_list>
<property_list>     ::= <property> | <property> <property_list>

<property>          ::= <indentation> <property_type> ":" <property_name>
                        [<restriction_block>]
<indentation>       ::= ":" | "::" | ":::" | "::::" | ":::::"
<property_type>     ::= "Attribute" | "Relation" | "Role" | "SetDo"

<restriction_block> ::= { "::" <restriction> }
<restriction>       ::= <restriction_type> ":" <restriction_value>
<restriction_type>  ::= "Condition" | "SetValue" | "Permission" | "Range" |
                        "Multiple" | "Required" | "ValueCondition" | "SetRange" |
                        "Default" | "Immutable" | "UniqueIdentifier" | "Unique" |
                        "DataType" | "SetDo"

# Expressions and queries

<expression>          ::= <js_expression> | <query_expression>
<query_expression> ::= <query_prefix> "(" <query_conditions> ")"
                       [<property_path>] [<array_access>]

<query_prefix>        ::= "$" | "$$"
<query_conditions>    ::= <query_condition> { "," <query_condition> }
<query_condition>     ::= <comparison_operator> | <logical_operator>
<comparison_operator> ::= <compare_op> "." <property_name> "(" <expression> ")"

<compare_op>          ::= "$EQ" | "$NE" | "$LT" | "$GT" | "$LE" | "$GE"
<logical_operator>    ::= "$OR" "(" <query_condition> { "," <query_condition> } ")"

<property_path>       ::= "." <property_name> { "." <property_name> }
<array_access>        ::= "[" <expression> "]"
```

## 5.3. Semantics of Queries and Expressions

BSL supports a system of variables for accessing the current state of the event graph:

- `$.property` — the value of the current individual's property (with the prefix `$.` execution fails if the value is missing; the `$$.` prefix is the safe version that returns `undefined`),
- `$Parent` — a reference to the parent property for nested events,
- `$CurrentActor` — the actor initiating the current event,
- `$CurrentIndividual` — the identifier of the current individual (whose event the engine is processing).

Queries allow for extracting data from the temporal graph, taking into account semantic connections:

```
# Names of all men over 18
$($EQ.$Model("Model Person"), $GT.age(18), $EQ.sex("man")).name

# Parties to all signed contracts
$($EQ.$Model("Model Contract"), $EQ.status("signed")).parties

# The last comment on a document
$($EQ.$Model("Model Document"), $EQ.$Id($.document)).comments[-1]
```

BSL is declarative, reactive, and verifiable; its syntax is readable and models run without compilation.

Thus, BSL represents a class of languages — executable specifications — where the boundary between describing requirements and their implementation is effectively erased.

# 6. Architecture of the boldsea-engine

The boldsea-engine is, in essence, a tool for semantic activity modeling, where the creation of business process models is implemented through the construction of a domain event ontology. The engine provides for: the interpretation and validation of semantic models, the processing of queries and expressions, the dataflow execution of business logic through a subscription mechanism, and the generation of system events (`CreateIndividual`, `EditIndividual`). The engine's architecture is designed to ensure reactivity, validation, and asynchronous business logic execution.

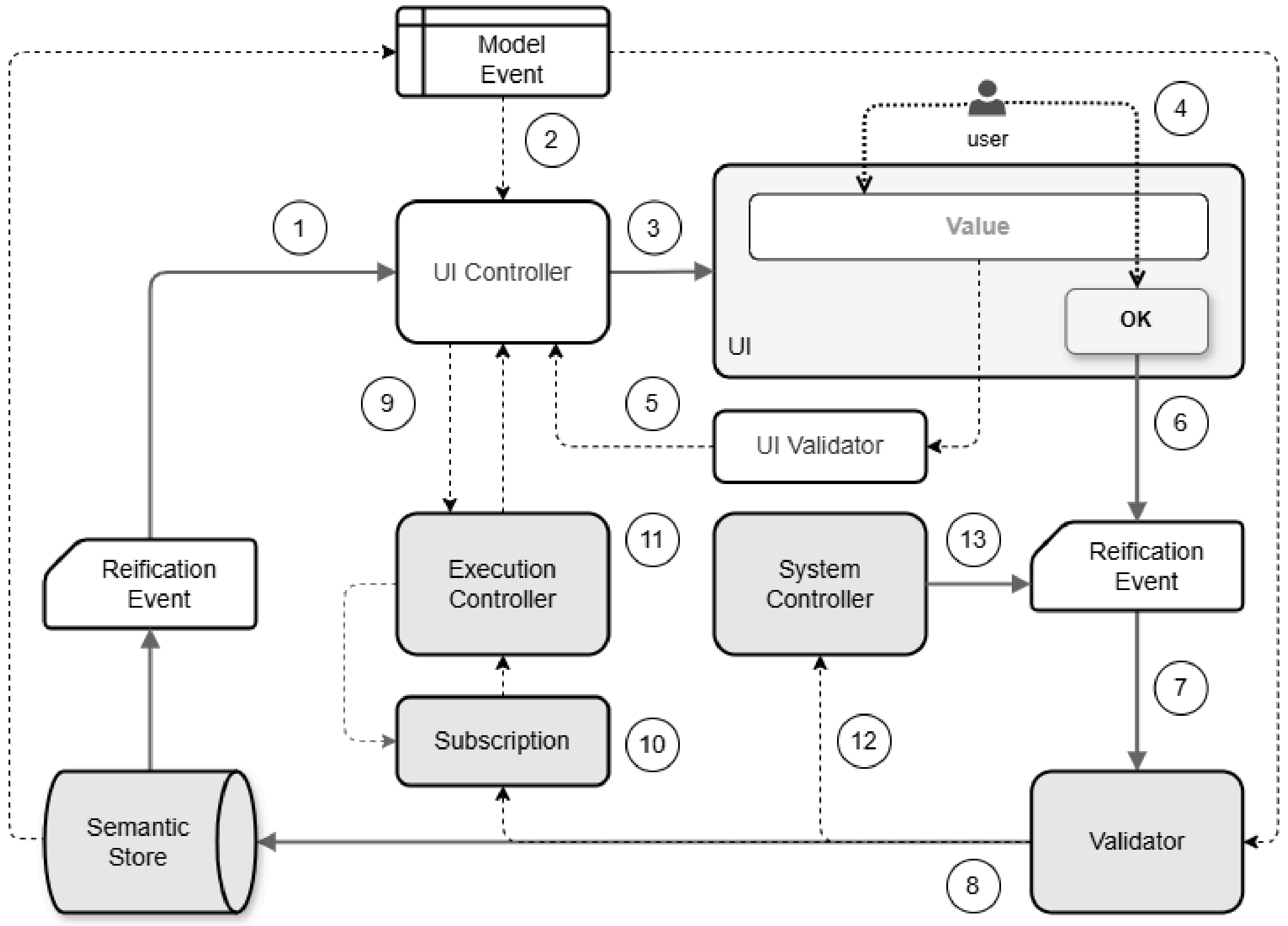

*Figure 5. Schematic diagram of the boldsea-engine architecture.*

The event processing workflow within the system comprises the following key steps:

1. *Initiation and Display:* Based on the existing state from the *Semantic Store* and the template from the *Model Event*, the UI Controller (2) forms and displays an interface for data entry to the user (3).
2. *Entry and Primary Validation:* The user enters a value (4), which undergoes primary validation in the *UI Validator* (5).
3. *Event Creation and Validation:* After confirmation (6), a new *Reification Event* is created and sent to the main *Validator* (7). The Validator checks the event for compliance with all semantic rules, restrictions, and access rights defined in the model, after which the event is saved to the store (8).
4. *Asynchronous Execution:* The saved event initiates a check in the *Execution Controller* (9). If the `Condition` is not met, the engine creates a lightweight subscription to changes in related events (10). As soon as the necessary events appear or are modified, the Execution Controller re-evaluates the condition and activates the model event (11).
5. *System events* (12), such as `SetDo`, are processed by the *System Controller* (13) for autonomous object management.

This architecture ensures a clear separation of tasks: the UI Controller is responsible for interaction, the Validator for data integrity, and the Execution and System Controllers for the asynchronous implementation of the dataflow logic.

# 7. Discussion and Limitations

The executable ontology architecture based on event semantics offers a new approach to modeling and automating business processes. However, like any innovative technology, it is associated with certain challenges and limitations.

## 7.1. Advantages of the Approach

A key advantage of event semantics is its architectural flexibility and adaptability. Unlike control-flow systems, where changing logic requires stopping and redeploying, the declarative nature of dataflow models, in most cases, allows them to be modified at runtime. This is critically important for dynamic environments where business rules can change "on the fly."

A second significant advantage is complete temporal transparency. Since all activity is recorded in the form of an immutable graph with cause-and-effect relationships between nodes, the system provides extensive opportunities for auditing, analysis, and process reproduction. Every fact in the system has a link to an actor and a timestamp, which creates a solid foundation for analytics and for training machine learning models on historical data.

Finally, the architecture ensures the unification of knowledge representation, data, and business logic. The single semantic format eliminates the need for complex integration layers between databases, rule engines, and ontologies (knowledge graphs). The use of common *vocabularies* creates the basis for genuine semantic interoperability between different applications and organizations.

## 7.2. Limitations and Challenges

Despite its significant advantages, adopting an architecture based on event semantics presents several challenges. The primary challenge is the cognitive barrier and the associated learning curve. Transitioning from the familiar object-oriented paradigm to the event-semantic paradigm requires analysts and developers to master a new methodology. However, this challenge can be mitigated by leveraging AI assistants for model creation, which can hide the underlying complexity of event semantics from the user.

Further research is needed on query performance against large temporal graphs. While load tests have not been conducted, modeling standard processes has not revealed any performance issues. Architectural solutions are being developed and tested to ensure data scaling and the optimization of complex temporal queries: semantic clustering by models, data indexing, the ability to archive inactive branches of the graph, and others.

It should also be noted that the proposed approach may be ineffective in certain areas: simple CRUD operations, high-frequency transactional systems, and cases where the event model is redundant.

And of course, the technology requires the development of an ecosystem of users and infrastructure. For mass adoption, debuggers and visualizers are needed to simplify work with executable ontologies. Also, to achieve true interoperability, it is necessary to create model repositories and standardized industry-specific vocabularies.

## 7.3. Directions for Future Research

The prospects for the technology's development lie in deepening its synthesis with advanced technologies. The most promising direction is deep integration with Large Language Models (LLMs). We expect LLMs to be used not only for generating and validating semantic models from natural language descriptions but also for leveraging the temporal graph as external, long-term semantic memory (both of which have already been implemented at the PoC level).

Another strategic direction is to use the boldsea-engine to build decentralized P2P networks that ensure the semantic clustering and interoperability of independent smart contracts. The event semantics architecture also provides all the necessary technological components for building flexible, transparent, and powerful multi-agent systems using: (1) the semantic graph as an environment for coordination and data exchange, (2) executable models to define agents, (3) a dataflow engine for their activation, and (4) LLM integration to create adaptive interfaces and on-the-fly generation of new actor models.

Finally, the formal basis of the technology opens the way for the development of automatic verification tools that can prove the correctness of business processes and the absence of logical conflicts, which is especially important for systems with high reliability requirements.

## 8. Conclusion

This paper presented a technology that implements the paradigm of executable ontologies for the unified modeling of complex dynamic systems. This approach successfully resolves the fundamental conflict between the static representation of knowledge and imperative process control by synthesizing the formal rigor of event semantics with the flexibility and parallelism of dataflow architecture.

Our contribution is threefold. First, it is a theoretical contribution in the form of the formalization of event semantics, which natively includes temporality, causality, and a subject-based approach, going beyond the limitations of traditional object ontologies (RDF/OWL). Second, it is an architectural innovation: the implementation of a system in which a semantic model, described in the BSL language, is executed directly by a dataflow engine without intermediate compilation into code. Third, the practical applicability of the approach was confirmed with examples of business processes, where measurable advantages in flexibility and development speed are achieved.

Future research will focus on scaling the approach for decentralized peer-to-peer (P2P) implementations and deepening its integration with artificial intelligence. We argue that event semantics is not merely a tool for creating more advanced applications but a foundation for a future digital environment where not isolated programs interact, but autonomous AI agents, through a shared, semantically rich, and cryptographically secure knowledge graph.

# Appendix

## 1. Glossary (Semantic Primitives)

- *Actor*: An entity that distinguishes, modifies, or acts. Formally, an actor is an authorized identifier. Through an authentication relationship, an actor can be linked to a human individual, as well as a robot, software agent, or sensor. Adding an actor to the event format allows for: (1) controlling the origin and validity of data, (2) obtaining an exhaustive description of actors as a set of events they generate, (3) including alternative opinions in the description of the domain.
- *Individual*: A unique entity that is distinguished, modified, or created by an Actor; an individual has spatial and/or temporal boundaries.
- *Concept*: the semantic type of an individual, what the actor associates the individual with (tree, document, person); to distinguish an individual means to specify which concept it belongs to; categories/classes of classifications are not concepts.
- *Action*: the semantic type of a time-distributed, ongoing individual, which is a partially ordered sequence of events.

- *Value*: what the actor distinguishes/changes in an individual (red, round, belongs to Anna); a value exists only as something distinguished on an individual; an individual is described as a set of distinguishable values.
- *Property*: the type of value (color, shape, ownership)
- *Type* (event type): the semantic type of an event, indicating the kind of change or fact being recorded. As a rule, an attribute, relation, or role is used as the `Type`.
  - *Attribute*: a property whose value an actor distinguishes on an individual directly, regardless of other individuals (color, shape); the value of an attribute has a fixed data type - `DataType` (string, integer, etc.) and can have a unit of measurement (`unit`).
  - *Relation*: a property whose value an actor fixes on an individual only in relation to another individual ("has part," "is a client"); the value of a Relation is an individual.
  - *Role*: the relation of an actor to an organization/project/community, which fixes the actor's access rights to individuals (their properties); the value of a role is an actor; formally, a role is a list of rights granted to actors; one actor can have different roles in different organizations.
- *Cause*: a field in an event containing links to one or more preceding events that served as the cause or condition for its generation. It forms the cause-and-effect relationships in the temporal graph.
- *Instance*: an event type that records the fact of declaring an individual of a certain concept.
- *Nested properties*: properties fixed on the properties of a concept (properties of properties); up to 5 levels of nesting are allowed; the properties on which nested properties are fixed are called parent properties.
- *SetDo*: a system act event that can be performed by the system automatically (without an actor's participation).
- *Model Event*: an event that records the presence of a certain property or act in a concept; a model event includes as nested events the restrictions on property values and the conditions for actualization (links to conditioning events).
- *Reification Event*: an event that instantiates a model event as a fact about an individual/action; it is created according to a model event, taking into account all restrictions and conditions.
- *Condition*: a condition that determines the possibility of creating a reification event based on a model event; it determines the ordering relationships between events; formally, it is defined by a logical expression composed of the values of the conditioning events.
- *Restriction (restricting property)*: a property of a model event that imposes restrictions on the possibility of creating a reification event and on its values (the number of reification events to be created, mandatory nature, data type, access rights, etc.).
- *Model*: an ordered list of model events describing a concept or an action. Individuals of concepts are created only according to models. The same individual can participate in different actions in which it can be represented by a different set of properties, so one concept can have several models (a medical person model, a person model in the HR department).

- *Vocabulary*: a named list of concepts or properties (attributes, relations, roles); used for importing/exporting a set of properties both autonomously and as part of applications.
- *Application*: a set of models and vocabularies combined to perform a fixed set of functions; an application is an element of import/export of functionality.
- *Organization*: a concept whose individual indicates a set of individuals of concepts, actions, and actors that implement purposeful activity within a fixed domain; the individual of the Organization concept is assigned applications according to which the organization's activity is implemented, roles, and other settings.

## 2. Integrated Development Environment (IDE)

The creation, testing, and execution of event models are carried out in an integrated development environment (IDE), which is essentially the administrative interface of the boldsea-engine. The environment provides a full set of tools for managing the entire lifecycle of executable ontologies.

Development begins in the *Vocabulary Editor*, where the basic semantic units — properties (attributes, relations, roles) — are created and combined into named, reusable vocabularies. These vocabularies serve as the basis for work in the *Model Editor*, which allows an analyst to build a tree of model events and declaratively assign them restricting properties that define the dataflow logic. The Editor also supports direct editing of models in the symbolic BSL notation with the possibility of connecting an LLM assistant to speed up development.

For debugging complex expressions and graph queries, the IDE contains a *Query Editor*. The execution and testing of models take place in the *Individual Editor*, which, based on the models, automatically generates an interface for creating concept instances and entering data, while ensuring compliance with all conditions, restrictions, and access rights.

Below is a screenshot of the Model Editor, which implements the practical example "Processing Product Request" discussed in the paper.

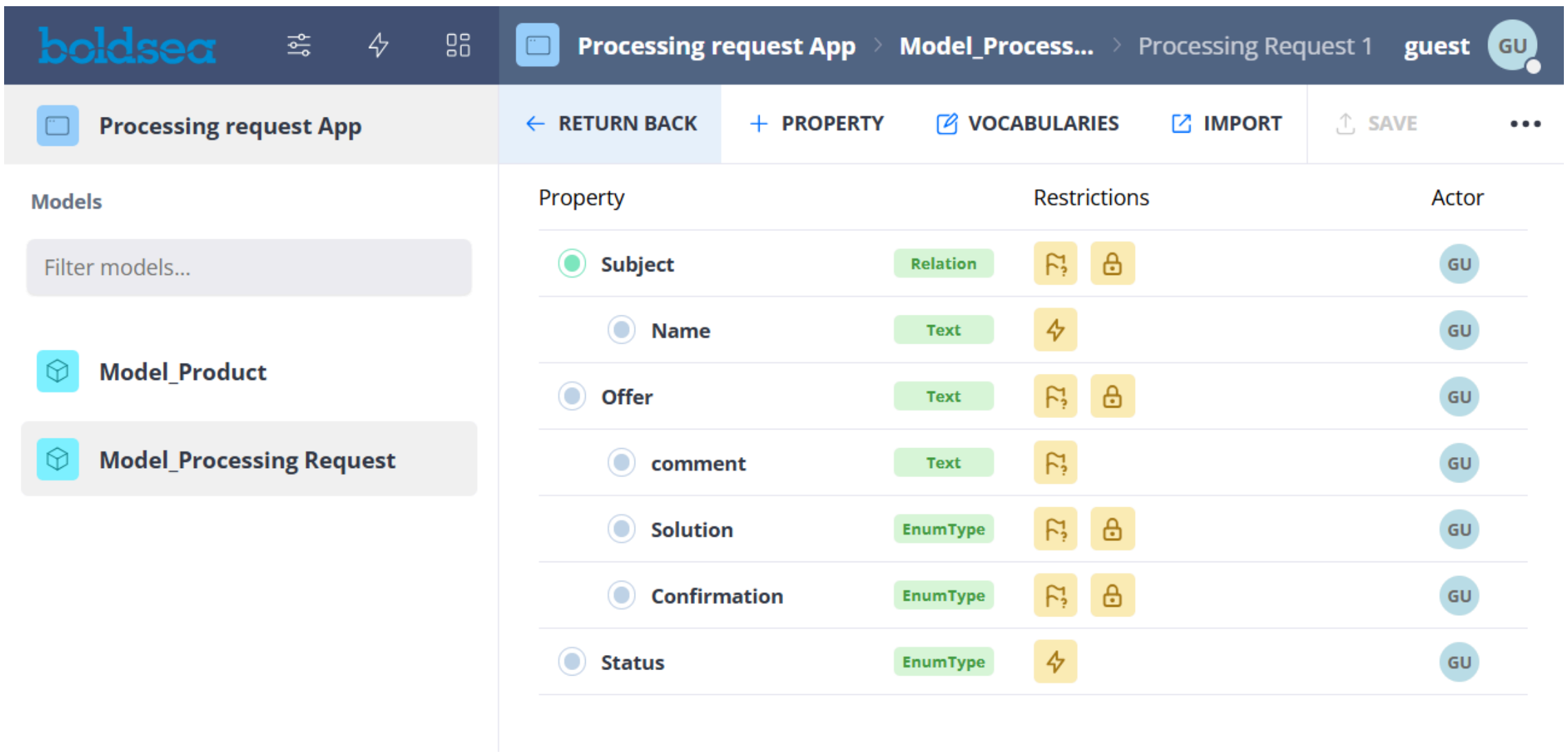

*Figure 6. boldsea IDE Model Editor with the "Processing Product Request" example.*

The IDE also contains a *UI controller* that creates user interface pages based on their description in the BSL language as individuals of the `View` concept using properties of a special vocabulary. That is, the interface is described as a special domain using the same tools as the business logic. View individuals (interface pages) are written to a special branch of the graph, separate from the business model branches. Such a use of BSL to describe the interface allows for using LLMs for interactive page customization and on-the-fly generation.

Thus, the IDE provides a complete no-code environment that allows business analysts to independently go through the entire process from the semantic design of a domain to the creation and debugging of executable business applications.

## 3. Current boldsea-engine Implementation

*Development Status.* As of September 2025, the engine is in a beta version. The interpretation of BSL models without compilation into code, the dataflow execution of business logic, and the cryptographic protection of the event graph based on hash chains have been implemented.

*Technological Architecture.* The engine and the integrated development environment (IDE) are written in TypeScript. The architecture allows the engine to be run in a browser without additional installation. Data storage is organized through in-memory structures with the ability to import/export a JSON dump.

*Functional Capabilities.* The IDE includes Model (with an AI assistant), Vocabulary, Query, and Individual Editors, with automatic generation of interfaces based on models. The export/import of BSL code is provided via the console. The UI controller supports the interpretation of basic view properties for building the user interface. The system supports authentication via cryptographic keys, a role-based access rights model, and provides full temporal traceability of changes.

*Proof of Concept.* The concepts of executable contracts, the transformation of BPMN diagrams into BSL models, and distributed voting processes have been tested. Load tests have not been conducted, but within the framework of the implemented PoC projects, no performance problems have been identified.

*AI Integration.* A basic integration with LLMs has been implemented for generating BSL models from natural language descriptions and parsing unstructured documents with the creation of their semantic models and individuals. A promising direction is the use of the event graph as external long-term memory for LLMs.

*Current Limitations and Prospects.* Current limitations include the need to optimize query performance for large graphs and to finalize the query language. Development plans include the creation of libraries of vocabularies, typical models, and support for compositional models for code reuse.